\documentclass[letterpaper, 10 pt, conference]{ieeeconf}  

\IEEEoverridecommandlockouts                              

\usepackage{graphicx}
\usepackage{threeparttable}
\usepackage{Dcolumn}
\usepackage{multirow}
\usepackage[table]{xcolor}
\usepackage{booktabs}
\usepackage{amsmath}
\usepackage{amssymb}
\usepackage{url}

\title{\LARGE \bf
BLVD: Building A Large-scale 5D Semantics Benchmark for Autonomous Driving
}

\author{Jianru Xue$^{1}$, Jianwu Fang$^{1,2}$, Tao Li$^{1}$, Bohua Zhang$^{1}$, Pu Zhang$^{1}$, Zhen Ye$^{1}$ and Jian Dou$^{1}$
\thanks{*This work was supported the National Key R\&D Program Project of China (No. 2016YFB1001004), National Natural Science Foundation of China (No. 61751308, 61773311 and 61603057), and China Postdoctoral Science Foundation (No. 2017M613152).}
\thanks{$^{1}$The authors are with the Institute of Artificial Intelligence and Robotics, Xi¡¯an Jiaotong University, Xi'an, China.
        {\tt\small jrxue@mail.xjtu.edu.cn}}%
\thanks{$^{2}$Jianwu Fang is also with the School of Electronic and Control Engineering, Chang'an University, Xi'an, China.
        {\tt\small j.w.fangit@gmail.com}}%
}

\begin{document}

\maketitle
\thispagestyle{empty}
\pagestyle{empty}

\begin{abstract}

In autonomous driving community, numerous benchmarks have been established to assist the tasks of 3D/2D object detection, stereo vision, semantic/instance segmentation. However, the more meaningful dynamic evolution of the surrounding objects of ego-vehicle is rarely exploited, and lacks a large-scale dataset platform. To address this, we introduce \texttt{BLVD}, a large-scale 5D semantics benchmark which does not concentrate on the static detection or semantic/instance segmentation tasks tackled adequately before. Instead, BLVD aims to provide a platform for the tasks of dynamic 4D (3D+temporal) tracking, 5D (4D+interactive) interactive event recognition and intention prediction. This benchmark will boost the deeper understanding of traffic scenes than ever before. We totally yield $249,129$ 3D annotations, $4,902$ independent individuals for tracking with the length of overall $214,922$ points, $6,004$ valid fragments for 5D interactive event recognition, and $4,900$ individuals for 5D intention prediction. These tasks are contained in four kinds of scenarios depending on the object density (low and high) and light conditions (daytime and nighttime). The benchmark can be downloaded from our project site \url{https://github.com/VCCIV/BLVD/}.
\end{abstract}

\section{INTRODUCTION}

Developing a safer, agiler and more dexterous autonomous vehicle with excellent ability of traffic scene understanding is the focus of much recent research in modern computer vision and robotic applications. Facing this urgent demand, many benchmarks have been constructed \cite{yin2017use} and the research progress is heavily linked with them. Two classic and attractive ones are KITTI Vision Benchmark Suite \cite{Geiger2013Vision} and Cityscapes \cite{Cordts2016The}, where KITTI was designed to test the functions of 2D/3D object detection, depth recovery, road segmentation, scene flow and optical flow, and Cityscapes were built to evaluate the semantic/instance segmentation performance, which derived the Citypersons \cite{DBLP:conf/cvpr/ZhangBS17} for video-level person detection.

\begin{figure}[!t]
	\centering
	\includegraphics[width=\linewidth]{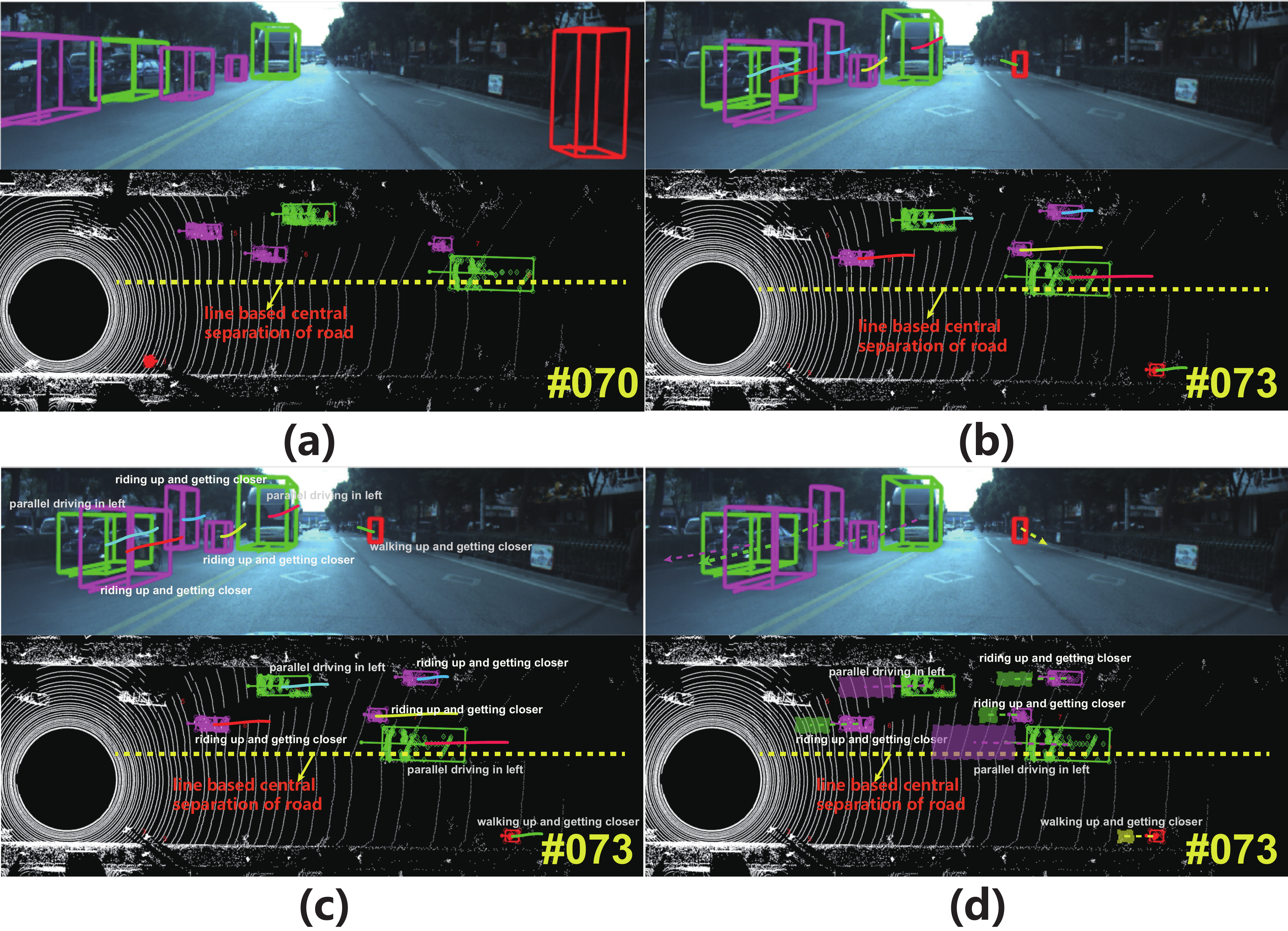}
	\caption{The task flow of BLVD with 5D semantic annotations. (a) denotes a frame with static 3D bounding boxes, (b) is another frame where the 4D trajectories of objects are shown, (c) specifies the frame where the 5D interactive event type is assigned in each trajectory, and (d) demonstrates the 5D intention prediction with the prediction of location, geometry structure of 3D bounding boxes, orientation and interactive event state.}
	\label{fig1}
\end{figure}

Although these benchmarks have pushed the autonomous driving research forward largely, they are all collected from day and sunny time and with limited scale. Some institutions aim to build larger and more challenging benchmarks. For example, Berkeley Deep Drive \cite{DBLP:journals/corr/abs-1805-04687} collected 100K color videos under various environments, different cities, diverse road conditions, where each video labeled 4 frames for every 10 seconds with instance label. Apollo scape \cite{DBLP:journals/corr/abs-1803-06184} built a benchmark with 147K images with fine instance annotation. Although these benchmarks are becoming larger and larger, many of them still concentrate on the static images or frames without continuous labeling, and do not label the large-scale 3D data \cite{DBLP:journals/corr/abs-1805-04687,DBLP:conf/iccv/NeuholdOBK17,DBLP:conf/cvpr/RosSMVL16}. In addition, the representation based on these vision datasets only can get static scene layout \cite{landsiedel2017road} and sparse moving pattern \cite{Geiger20143D} of the observed scene. However, the more meaningful 3D dynamic evolution of the surrounding objects of ego-vehicle when driving was rarely exploited, and lacks a large-scale platform with unifying metrics for performance evaluation~\cite{Xue2018A,Yao2017On,Ernst2016Behaviour,Schneemann2016Context}.

In this paper, we \textbf{b}uild a \textbf{l}arge-scale \textbf{5D} semantics benchmark (BLVD), specifically tailored on the tasks of 4D (3D+temporal) tracking, 5D (4D+interactive) interactive event recognition and intention prediction in autonomous driving. We defined three kinds of participants, including \emph{vehicles}, \emph{pedestrians} and \emph{riders}, where riders contains cyclists and motorbikes which always demonstrate ruleless moving. The benchmark is constructed by a self-driving platform providing multiple kinds of sensors for surrounding perception, including a Velodyne HDL-64E LIDAR scanner, a GPS/inertial system, two multi-view cameras with high resolution. It is worth noting that all the sensors are registered and synchronized automatically.
Different from many other datasets, this benchmark is collected under different driving scenarios (urban and highway), various light conditions (daytime and nighttime), and with full 5D semantics annotation for 120K frames. BLVD benchmark exceeds the previous efforts to deeper traffic scene understanding, in terms of annotation size, light richness, and tasks. In particular, the large-scale 4D participant trajectories and 5D interaction between surrounding objects with ego-vehicle are defined in the vehicle-based coordinate system. One typical snapshot representing a task flow from static 3D annotations to 5D intention prediction of our benchmark is shown in Fig. \ref{fig1}.

\section{BLVD benchmark}
\subsection{Sensors and Acquisitions}
Our dataset aims at deeper understanding for the dynamic traffic scene. Multiple kinds of sensors are equipped for surrounding perception, including a Velodyne HDL-64E LIDAR scanner (10Hz, 64 laser beams, range of 100m), a GPS/inertial system, two multi-view color cameras with high resolution (30Hz, resolution: $1920\times500$ pixels larger than $1242\times375$ of KITTI \cite{Geiger2013Vision}). The sensors are mounted on the top of our vehicle, where the cameras are built-in a box with a windshield and capture the front view of road. Velodyne HDL-64E unit can provide accurate 3D information from moving platforms. The egomotion in the 3D laser measurements is compensated by our GPS/IMU system. Different from the data collection way from multiple cities (e.g., Cityscapes~\cite{DBLP:conf/cvpr/ZhangBS17}), we gather the data in a same city (Changshu, Jiangsu province, China) under different light conditions (daytime and nighttime), diverse densities (low and high density), and distinct scenarios (highway and urban), where the dynamic participants are fully annotated in 3D mode. In this benchmark, we design a fast online calibration method which can efficiently register and synchronize camera and 3D LIDAR, and costs only one day for the accurate calibration of 120k frames. Consequently, we obtain $654$ calibrated video clips with the frequency of 10Hz, which contain images and 3D point cloud simultaneously.
\subsection{5D Semantics Specifications}

In the annotation process, we locate in vehicle-based coordinates system. Unlike most of benchmarks relying online crowd-sourcing labeling, we hired a set of annotators to assign 5D semantics for each individual, and frequently gathered the feedbacks of each annotator to form many consistent and reasonable labeling principles.

Our benchmark aims to assist the ego-vehicle to comprehend its driving scene and make effective decision. Our benchmark removed the perception range which is not helpful for driving decision of ego-vehicle, such as the zones outside a guardrail. This strategy is different from other detection benchmarks which labeled all participants in the image. Similar to KITTI, we annotated the participants that appear both in images and 3D point clouds within 50 meters in the front view. We developed a semi-automatic tool for fast annotation, where the corresponding image and its bird's eye view (BEV) of 3D point clouds were shown in the interface.
\begin{figure}[!t]
\centering
\includegraphics[width=\linewidth]{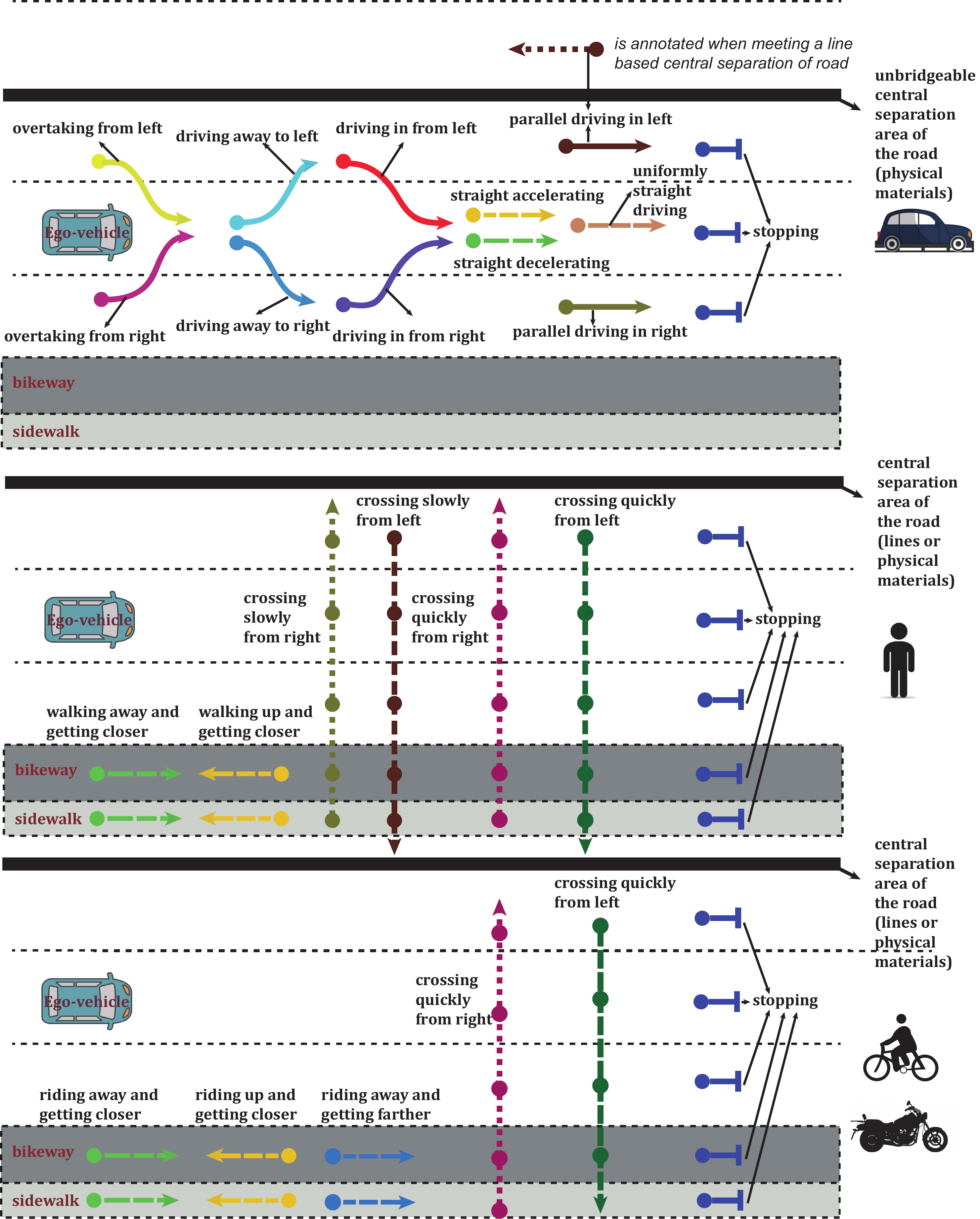}
    \caption{The illustration of interactive events standing at vehicle-based coordinate system. From top to bottom, the event types of vehicles, pedestrians and riders are demonstrated. Note that, there is an extra event type of participants (specified as ``\emph{others}") for denoting the ambiguous interactive event.}
\label{fig2}
\end{figure}
\begin{figure}[!t]
\centering
\includegraphics[width=0.7\linewidth]{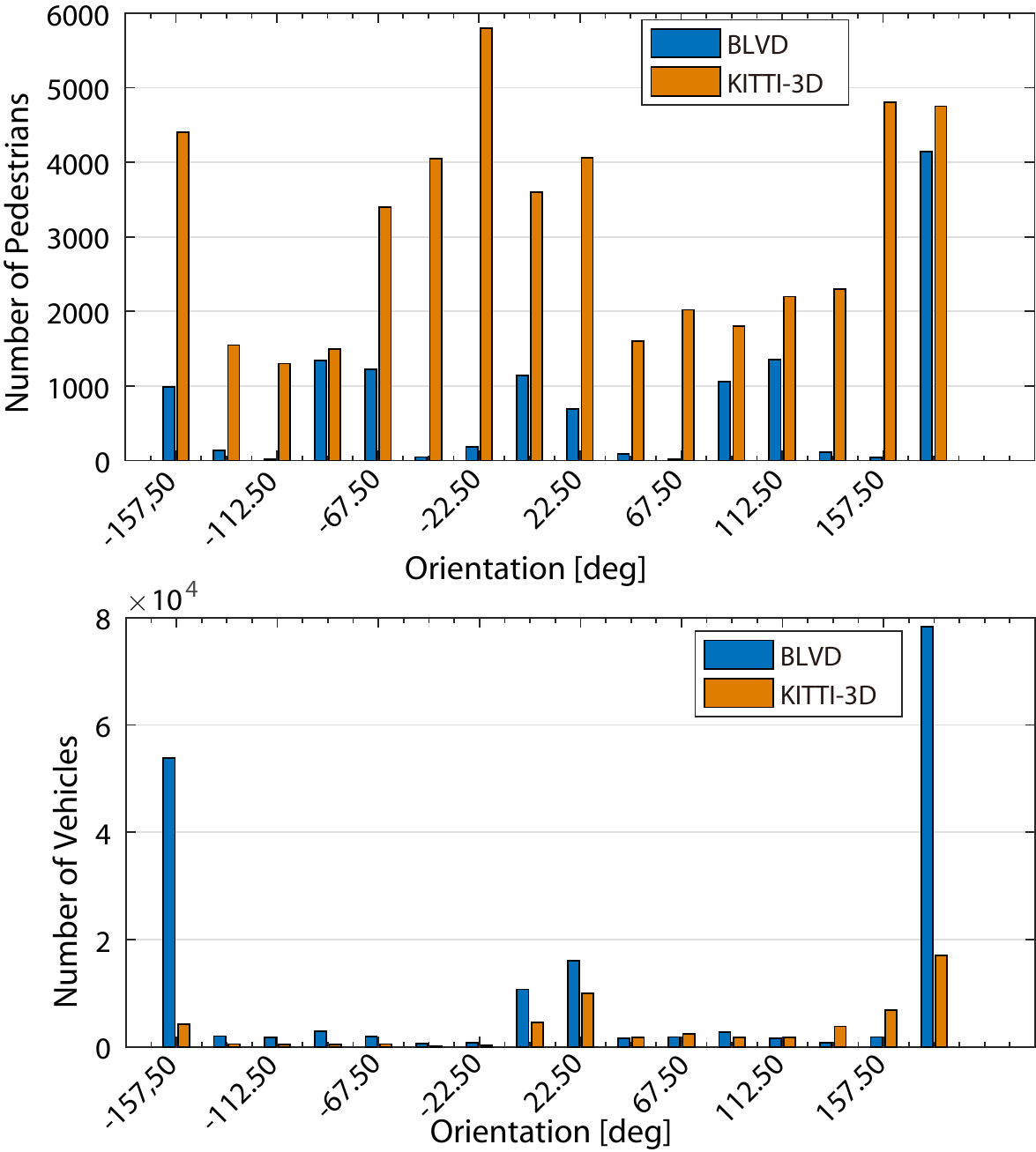}
    \caption{The orientation distribution of BLVD and KITTI-3D~\cite{Geiger2013Vision}.}
\label{fig3}
\vspace{-1.5em}
\end{figure}
\begin{figure}[htbp]
\centering
\includegraphics[width=\linewidth]{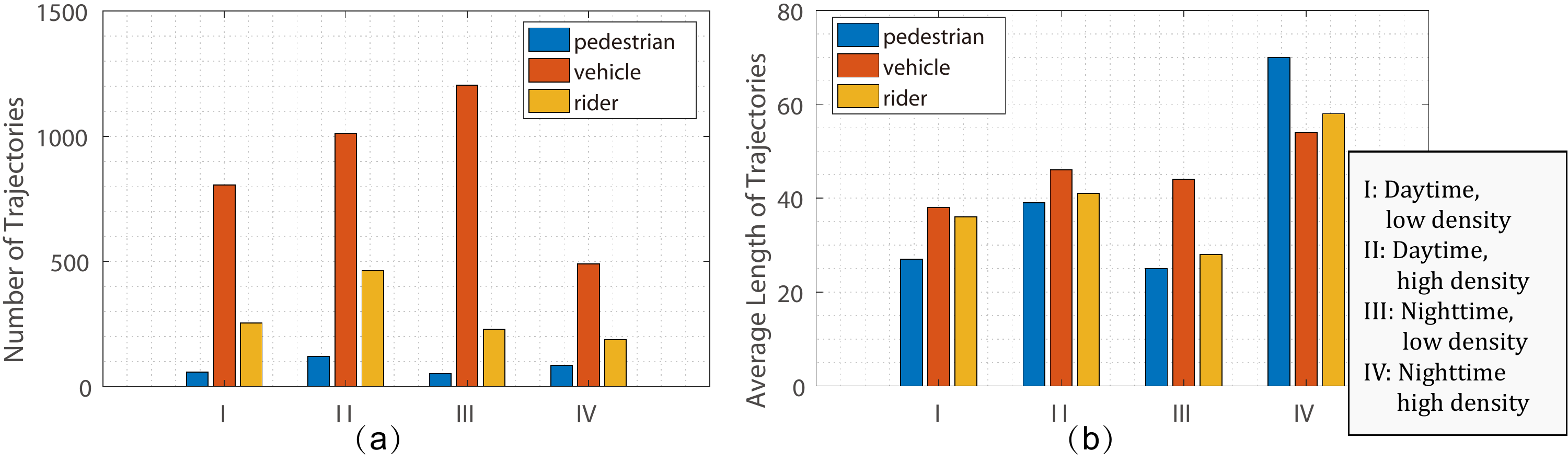}
    \caption{The statistics of trajectories. (a) denotes the number of individuals under four different scene conditions and (b) is the average trajectory length.}
\label{fig4}
\vspace{-1.5em}
\end{figure}
\begin{table*}[htbp]\small
	\centering
	\begin{threeparttable}
		\caption{Training and testing set statistic w.r.t., NI (number of individuals) and TTL (total trajectory length) under four kinds of scene conditions of I (Daytime with low density), II (Daytime with high density), III (Nighttime with low density) and IV (Nighttime with high density).}
		\label{tab:performance_comparison}
		\begin{tabular}{cccccccccccc}
			\toprule
			\multirow{2}{*}{Classes}&\multirow{2}{*}{Data Splits}&
			\multicolumn{2}{c}{I}&\multicolumn{2}{c}{II}&\multicolumn{2}{c}{III}&\multicolumn{2}{c}{IV}&\multicolumn{2}{c}{Total}\cr
			\cmidrule(lr){3-4} \cmidrule(lr){5-6}\cmidrule(lr){7-8}\cmidrule(lr){9-10}\cmidrule(lr){11-12}
			&&NI&TTL&NI&TTL&NI&TTL&NI&TTL&NI&TTL\cr
			\midrule
            \multirow{2}{*}{\emph{Pedestrians}}&
			Training&28&819&61&2,890&30&849&41&4,154&160&8,172\cr
            &Testing&30&790&60&1,808&23&517&44&1,838&157&4,953\cr

            \multirow{2}{*}{\emph{Vehicles}}&
			Training&433&16,114&546&24,795&545&23,423&215&17,209&1,739&81,541\cr
	        &Testing&355&14,242&442&21,162&659&30,069&276&9,533&1,732&75,006\cr

            \multirow{2}{*}{\emph{Riders}}&
            Training&121&4,790&250&10,197&113&3,272&79&5669&563&23,928\cr
             &Testing&122&4,004&205&8,349&116&3,242&108&5,187&551&20,782\cr
         \bottomrule
		\end{tabular}
	\end{threeparttable}
\end{table*}

\textbf{3D bounding box:} Each 3D bounding box is labeled by manually click four points (top-left point, top-right point, bottom-right point and an arbitrary point for determining orientation) on the BEV of 3D point clouds shown in screen. After these operations, the length, width and orientation of a 3D bounding box is automatically computed and stored. The height of the bounding box is automatically determined by finding the minimum value enclosing most of the 3D points in the 3D box. Then, we project the 3D point cloud back to image plane and visualize it for a double check. All the labeled 3D annotations can be easily modified for guaranteing accuracy. The benchmark is fully annotated, which yields $214,922$ 3D annotations comprising $179,073$ vehicles, $12,599$ pedestrians, and $51,917$ riders. We compare BLVD with the ones of KITTI in terms of orientation distribution of vehicles and pedestrians, as shown in Fig. \ref{fig3}. The vehicle annotations in BLVD has more complexity than KITTI-3D. BLVD benchmark aims to provide a platform for deeper traffic scene understanding, we focused on the individuals which may interact with the ego-vehicle. Therefore, we did not care the pedestrians outside the useful perception range.

\textbf{4D object IDs:} When assigning the IDs for each individual, we adopt the principle that the first frame must be labeled carefully as much as possible, where the 3D bounding boxes, orientations, IDs and object classes are accurately initialized. Then, for subsequent frames, we only need to drag each box to its new location with a tiny modification of orientation and interactive event state. This strategy can largely boost the annotation efficiency. We assign the same ID for the object which has appeared before and re-appears again in the same video clip, and add new ID for newly observed objects. The IDs increase till the end of the video clip, and are re-assigned for a new clip from 1. We obtain $4,902$ valid individuals with the trajectory length of overall $214,922$ points.

\textbf{5D interactive event:} In driving, perceiving the interactive events of other participants to ego-vehicle is necessary for making a reasonable decision. We defined $13$ kinds of events, $8$ kinds of events and $7$ kinds of events for vehicles, pedestrians and riders, respectively. For a clearer understanding, the types of events are demonstrated in Fig. \ref{fig2}. Note that, we assign these event types to each point of trajectories. Each kind of event here corresponds its reasonable road location that the participant may appear and is determined by multiple trajectory points. Additionally, this benchmark also labeled $8$ kinds of events of ego-vehicle. They are denoted as: \emph{straight accelerating}, \emph{straight decelerating}, \emph{turning right}, \emph{turning left}, \emph{uniformly straight driving}, \emph{changing line to left}, \emph{changing line to right}, and \emph{stopping}. We obtain $6,004$ valid event fragments of surrounding participants.

\textbf{5D intention:} 5D intention prediction inherits the 4D trajectories. Different from the location based intention works \cite{Alahi2016Social,DBLP:journals/corr/abs-1710-04689}, we advocate a prediction of locations, event types, geometrical structures of 3D bounding boxes, and orientations.

\subsection{Dataset Splits}
We split the data as training and testing sets, involving a balanced distribution regarding following properties of equal shares: 1) light conditions (daytime and nighttime), 2) participants densities (low and high), 3) participant category (pedestrians, vehicles and riders), and 4) event types. The statistics of BLVD will be demonstrated in following section.

\section{Benchmarking}
There is no publicly available dataset like BLVD involving 4D tracking, 5D interactive event recognition, and 5D intention prediction simultaneously. We will analyze the statistics of BLVD and compare it with other benchmarks when focusing certain tasks. Additionally, we provide the metrics for the performance evaluation of each task.

\subsection{4D Tracking}
The first task is to track multiple 3D individuals in one video clip. In the literatures, the works for multi-object tracking mainly are based on RGB videos \cite{DBLP:journals/corr/abs-1802-09298}, stereo sequential images \cite{romero2016variational} or LiDAR point sequence \cite{DBLP:journals/ras/WangWLMY17}. Recently, Frossard and Urtasun \cite{DBLP:journals/corr/abs-1806-11534} addressed the object tracking by fusing RGB image and 3D point cloud on KITTI benchmark which is only with $40$ sequences for 4D tracking. We contribute a larger one owning $654$ video sequences. In our benchmark, we separate the trajectories as the ones of vehicles, pedestrians and riders.

\textbf{Statistic analysis:} A trajectory is valid when its length is larger than $10$ frames whatever the participant is. The statistics of trajectories with respect to lighting conditions, participant densities and classes are analyzed in Fig. \ref{fig4}. With the condition of participant density and light, we gathered $164$, $283$, $114$, $93$ video clips under daytime with low density, nighttime with low density, daytime with high density and nighttime with high density, respectively. The corresponding average numbers of individuals are $10.7$, $5.3$, $14.0$ and $8.2$, respectively. Because the number of video clips with low density is larger than the ones of high density, the videos of low density have more individuals than the ones of high density. To depress this imbalance, we make an equal distribution for data split. The statistics for training set and testing set are listed in Table. I.

\textbf{Metrics:} To assess the performance of tracking hypothesis, we rely on the metrics evaluating both accuracy and efficiency. The common known metrics for multiple object tracking are launched by MOT challenge\footnote{https://motchallenge.net/}. For 3D object tracking, multiple object tracking accuracy (MOTA)~\cite{Bernardin2008Evaluating}, multiple object tracking precision (MOTP)~\cite{Bernardin2008Evaluating}, ID F1 Score (IDF1)~\cite{Ristani2016Performance}, mostly tracked targets (MT), mostly lost targets (ML), total number of false positives (FP), total number of missed targets (FN),  total number of identity switches (ID Sw)~\cite{Li2009Learning}, total number of times of trajectory fragmenting (Frag) and processing speed (Hz) all can be used, where the most important metrics for evaluating accuracy are MOTA and MOTP. MOTA combines three error sources of false positives, missed targets and identity switches, and MOTP measures the misalignment between the annotated and the predicted bounding boxes. The detailed meaning of each metric can be referred to the official website of 3D tracking of MOT challenge.

\subsection{5D Interactive Event Recognition}
For understanding the dynamic evolution of traffic scene, event recognition or reasoning is a core problem, because it can reflect the dynamic evolution process of scene with tractable reasoning strategy \cite{Waltz1980Understanding}. As far as we know, we are the first attempt to launch the 5D interactive event recognition platform. Accurate interactive event recognition can supply a powerful information for path planning and motion decision.
\begin{figure}[htbp]
\centering
\includegraphics[width=0.75\linewidth]{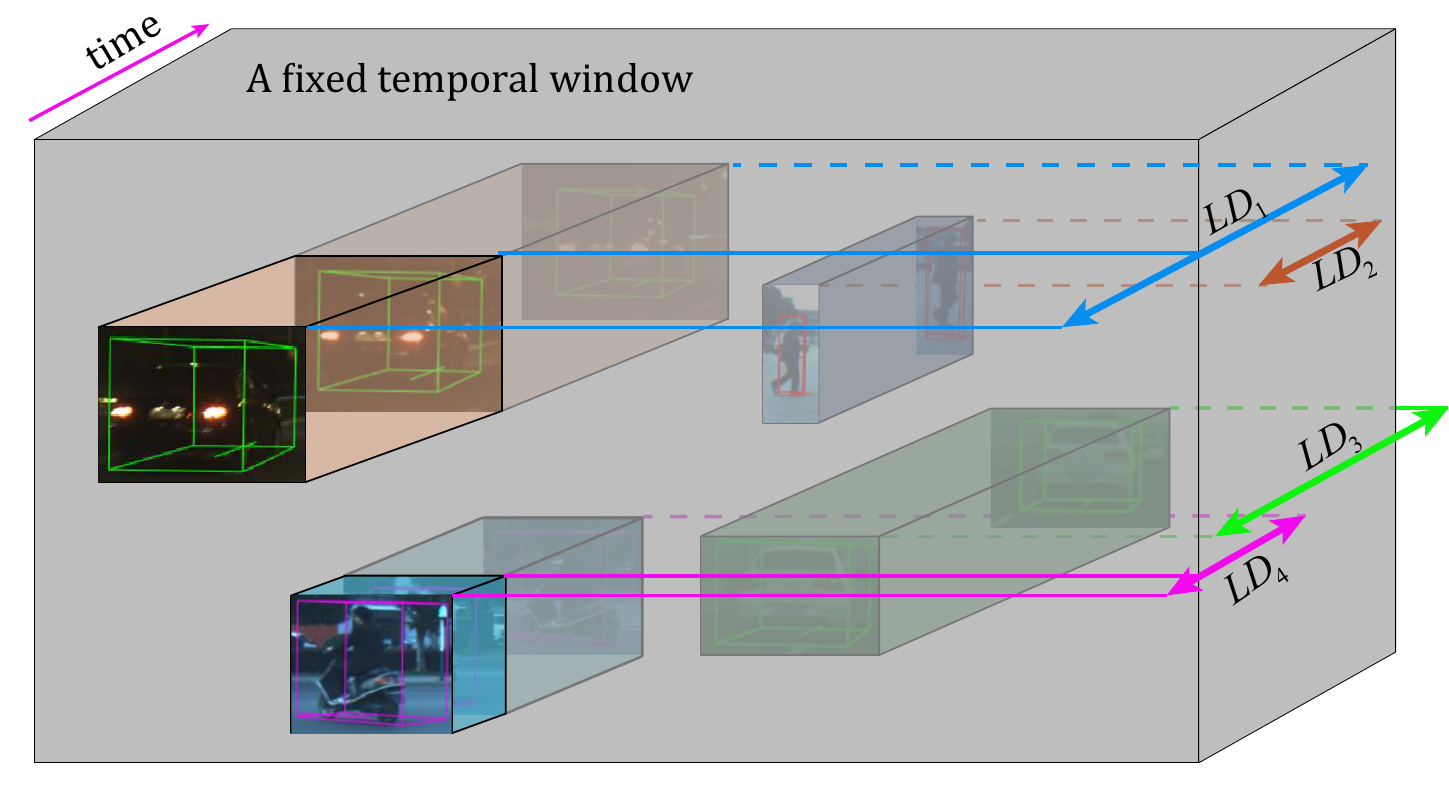}
    \caption{The illustration for individuals with different length of duration. $LD_1$ and $LD_4$ links some frames before entering a fixed temporal window, $LD_2$ links the frames falling into the fixed temporal window, and $LD_3$ associates some frames exceeding the fixed temporal window.}
    \vspace{-1.0em}
\label{fig4}
\end{figure}
\begin{figure}[htbp]
\centering
\includegraphics[width=\linewidth]{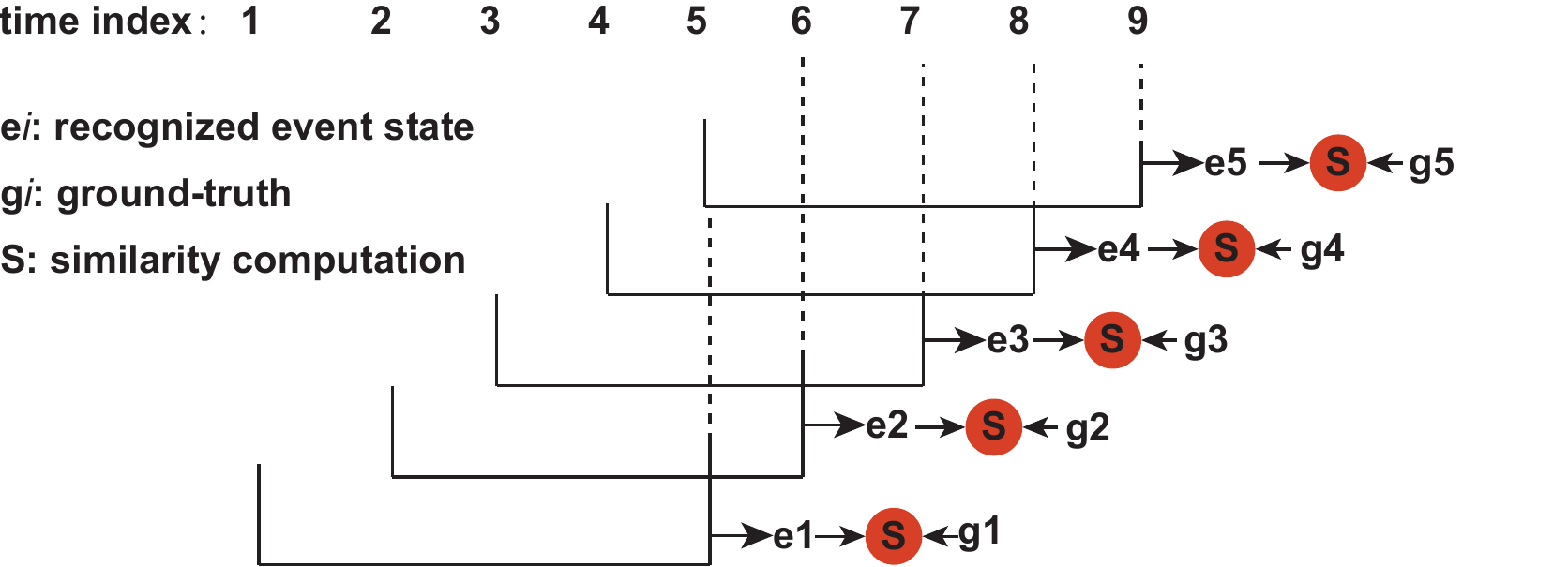}
    \caption{A schematic example for recognizing an event fragment with 5 nodes of state chain.}
\label{fig5}
\end{figure}

\begin{figure}[htbp]
\centering
\includegraphics[width=0.9\linewidth]{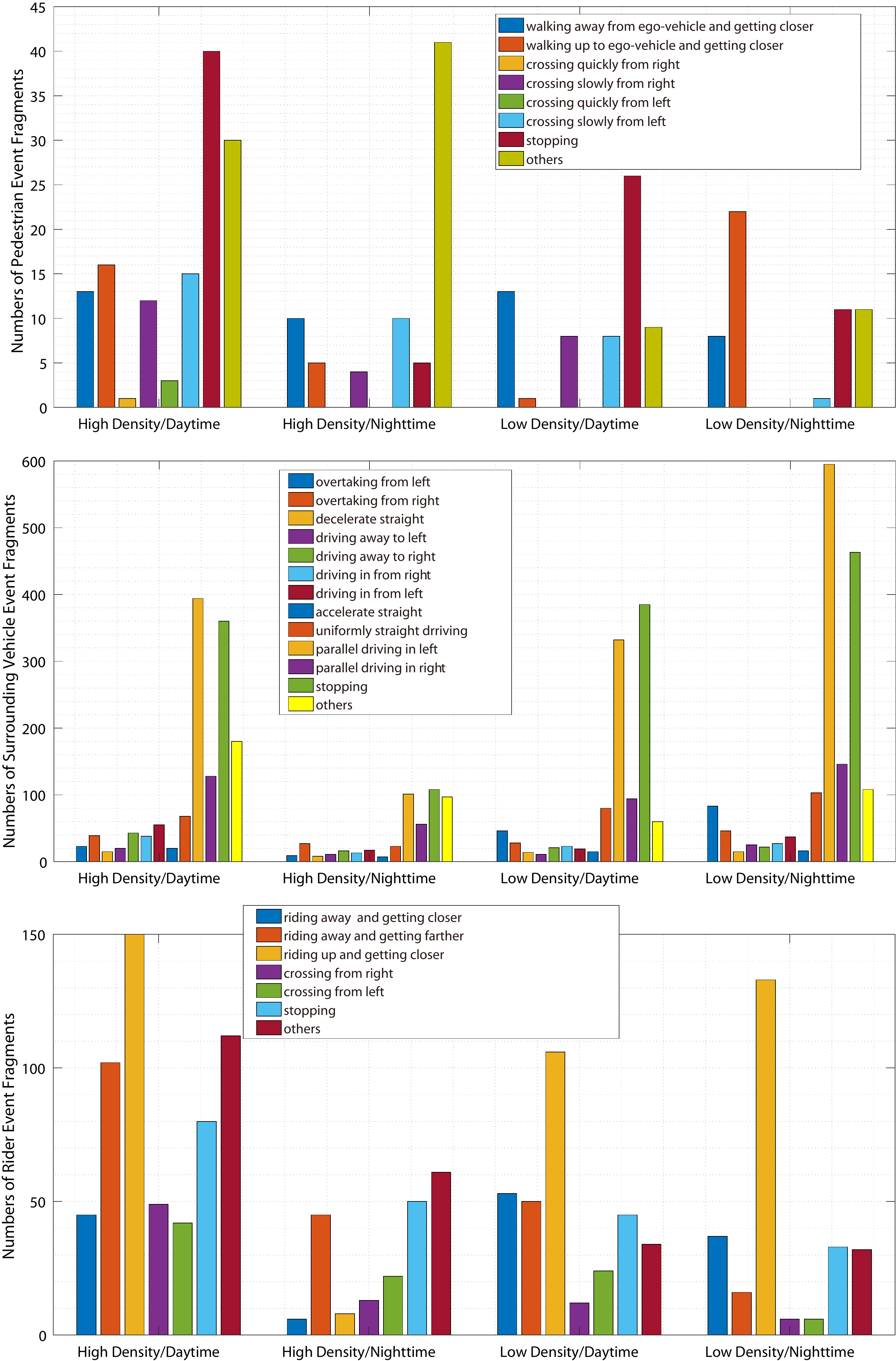}
    \caption{Statistics of interactive event type of pedestrians, vehicles and riders in terms of light condition and participant density.}
    \vspace{-1.5em}
\label{fig6}
\end{figure}

\begin{figure*}[htbp]
\centering
\includegraphics[width=0.9\linewidth]{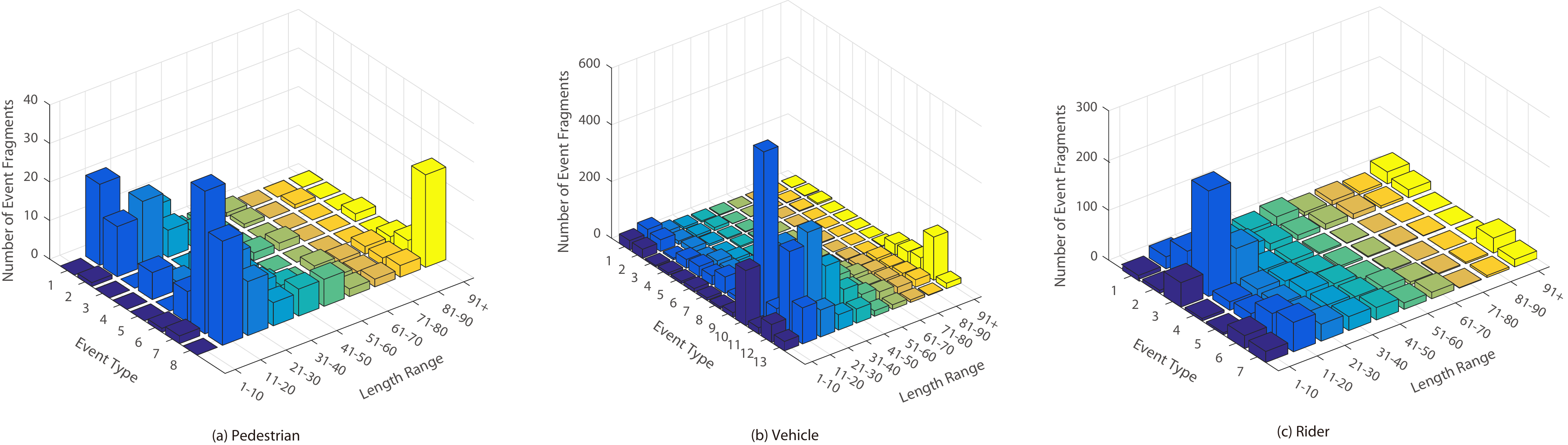}
    \caption{Statistics of the number of event fragments with respect to event type and and length range of fragments. The event type indexes in axis are the same as ones in Table. II.}
\label{fig8}
\end{figure*}
\textbf{Statistic analysis:} As aforementioned, we totally obtain $4902$ individuals with valid trajectories. We assigned interactive event type into each point of trajectories. Actually, each kind of interactive event has different temporal durations, as shown by the example in Fig. \ref{fig4} with a reference of a fixed temporal window. In one trajectory, there may be multiple kinds of interactive events.

Apparently, it is difficult to recognize the interactive events of all the individuals within a fixed temporal window. Therefore, this benchmark paves this task as recognizing the \emph{interactive event state} in each point lying on individuals' trajectories, and formulates it as a \textbf{sequence to sequence recognition} problem. In this problem, the recognition of an event state relies on multiple nearly observed trajectory points. Hence, the input sequence should be longer than the output one. The detailed illustration is demonstrated in Fig. \ref{fig5}. Under this problem setting, we fix the minimum number of points for recognizing the first point of output sequence as $5$, and the minimum number of points in output sequence as $3$. Consequently, the trajectories with the length larger than $8$ are valid for this task, and yield $6,004$ fragments for interactive event recognition. The statistics of event types in terms of participant category, light condition and participant density are demonstrated in Fig. \ref{fig6}. Additionally, we also analyzed the number of event fragments in relative to the event type and and length range of each fragment. The analyzed statistics are shown in Fig. \ref{fig8}. In this benchmark, the splitting of event recognition data inherits the principle of the one of trajectories, i.e, the training set and testing set of interactive event are taken out from the training and testing trajectories, as analyzed in Table. I.
\begin{table*}\scriptsize
\centering
	\begin{threeparttable}
		\caption{Performance matrix for interactive event recognition of vehicles, pedestrians and riders under four kinds of scene conditions of I (Daytime with low density), II (Daytime with high density), III (Nighttime with low density) and IV (Nighttime with high density).}
		\label{tab:performance_comparison}
		\begin{tabular}{ccccc|cc}
			\toprule
          Interactive Event Type&I&II&III&IV&Precision&Recall\cr
          \hline
            vehicle overtaking from left&\emph{TP/FP/FN}&\emph{TP/FP/FN}&\emph{TP/FP/FN}&\emph{TP/FP/FN}&\emph{P}&\emph{R}\cr
            vehicle overtaking from right&\emph{TP/FP/FN}&\emph{TP/FP/FN}&\emph{TP/FP/FN}&\emph{TP/FP/FN}&\emph{P}&\emph{R}\cr
            vehicle driving away to left&\emph{TP/FP/FN}&\emph{TP/FP/FN}&\emph{TP/FP/FN}&\emph{TP/FP/FN}&\emph{P}&\emph{R}\cr
            vehicle driving away to right&\emph{TP/FP/FN}&T\emph{P/FP/FN}&\emph{TP/FP/FN}&\emph{TP/FP/FN}&\emph{P}&\emph{R}\cr
            vehicle driving in from left&\emph{TP/FP/FN}&\emph{TP/FP/FN}&\emph{TP/FP/FN}&\emph{TP/FP/FN}&\emph{P}&\emph{R}\cr
            vehicle driving in from right&\emph{TP/FP/FN}&\emph{TP/FP/FN}&\emph{TP/FP/FN}&\emph{TP/FP/FN}&\emph{P}&\emph{R}\cr
            vehicle parallel driving in left&\emph{TP/FP/FN}&\emph{TP/FP/FN}&\emph{TP/FP/FN}&\emph{TP/FP/FN}&\emph{P}&\emph{R}\cr
            vehicle parallel driving in right&\emph{TP/FP/FN}&\emph{TP/FP/FN}&\emph{TP/FP/FN}&\emph{TP/FP/FN}&\emph{P}&\emph{R}\cr
            vehicle straight accelerating&\emph{TP/FP/FN}&\emph{TP/FP/FN}&\emph{TP/FP/FN}&\emph{TP/FP/FN}&\emph{P}&\emph{R}\cr
            vehicle straight decelerating&\emph{TP/FP/FN}&\emph{TP/FP/FN}&\emph{TP/FP/FN}&\emph{TP/FP/FN}&\emph{P}&\emph{R}\cr
            vehicle uniformly straight driving&\emph{TP/FP/FN}&\emph{TP/FP/FN}&\emph{TP/FP/FN}&\emph{TP/FP/FN}&\emph{P}&\emph{R}\cr
            vehicle stopping&\emph{TP/FP/FN}&\emph{TP/FP/FN}&\emph{TP/FP/FN}&\emph{TP/FP/FN}&\emph{P}&\emph{R}\cr
            walking/riding away and getting closer&\emph{TP/FP/FN}&\emph{TP/FP/FN}&\emph{TP/FP/FN}&\emph{TP/FP/FN}&\emph{P}&\emph{R}\cr
            walking/riding up and getting closer&\emph{TP/FP/FN}&\emph{TP/FP/FN}&\emph{TP/FP/FN}&\emph{TP/FP/FN}&\emph{P}&\emph{R}\cr
            riding away and getting farther&\emph{TP/FP/FN}&\emph{TP/FP/FN}&\emph{TP/FP/FN}&\emph{TP/FP/FN}&\emph{P}&\emph{R}\cr
            pedestrian/rider crossing slowly from right&\emph{TP/FP/FN}&\emph{TP/FP/FN}&\emph{TP/FP/FN}&\emph{TP/FP/FN}&\emph{P}&\emph{R}\cr
            pedestrian/rider crossing slowly from left&\emph{TP/FP/FN}&\emph{TP/FP/FN}&\emph{TP/FP/FN}&\emph{TP/FP/FN}&\emph{P}&\emph{R}\cr
            pedestrian crossing quickly from right&\emph{TP/FP/FN}&\emph{TP/FP/FN}&\emph{TP/FP/FN}&\emph{TP/FP/FN}&\emph{P}&\emph{R}\cr
            pedestrian crossing quickly from left&\emph{TP/FP/FN}&\emph{TP/FP/FN}&\emph{TP/FP/FN}&\emph{TP/FP/FN}&\emph{P}&\emph{R}\cr
            pedestrian/rider stopping&\emph{TP/FP/FN}&\emph{TP/FP/FN}&\emph{TP/FP/FN}&\emph{TP/FP/FN}&\emph{P}&\emph{R}\cr
            others (vehicles/pedestrians/riders)&\emph{TP/FP/FN}&\emph{TP/FP/FN}&\emph{TP/FP/FN}&\emph{TP/FP/FN}&\emph{P}&\emph{R}\cr
            \hline
            Precision&\emph{P}&\emph{P}& \emph{P}&\emph{P}& & \cr
            Recall&\emph{R}&\emph{R}&\emph{R}&\emph{R}& & \cr
			\bottomrule
		\end{tabular}
	\end{threeparttable}
\label{tab2}
\end{table*}

\textbf{Metrics:} We formulate the 5D interactive event recognition task as a sequence to sequence recognition problem. The performance evaluation depends on the precision $P=TP/(TP+FP)$ and recall $R=TP/(TP+FN)$ of the recognized event states, where $TP$ is true positives, $FP$ is false positives, and $FN$ is false negatives. These metrics are also adopted by \cite{Kristoffersen2016Towards,Dueholm2016Trajectories} for trajectory-based event recognition. Specially, the performance evaluation of our benchmark will be carried out on different situations, including low-density/daytime, high-density/daytime, low-density/nighttime and high-density/nighttime, and can be clearly visualized by filling out Table. II.

\subsection{5D Intention Prediction}
In autonomous driving, the purpose of aforementioned 4D object tracking and 5D interactive event recognition is to provide an accurate reasoning clue for future driving, i.e., to make a precise movement prediction of surrounding participants in the future. Thus, ego-vehicle can smoothly pass the observed scene and reach the destination as fast as possible. 5D intention prediction is another task launched for the first time by our benchmark. Previous intention works \cite{Alahi2016Social,DBLP:journals/corr/abs-1710-04689} in computer vision concentrate on predicting the future time-step locations of the target. Commonly, they evaluate the performance by two metrics proposed in \cite{Alahi2016Social}: average displacement error (ADE) and final displacement error (FDE), where ADE denotes the Euclidean distance between the predicted trajectory and the actual trajectory averaged over all time-steps for all targets, and FDE specifies the average Euclidean distance between the predicted trajectory point and the actual trajectory point at the end of $n$ time steps. ADE and FDE are computed as:
\begin{equation}\footnotesize
\begin{gathered}
  ADE = \frac{{\sum\nolimits_{i = 1}^N {\sum\nolimits_{m = 1}^M {\sum\nolimits_{t = {t_{obs}} + 1}^{{t_{obs}} + {t_{pred}}} {\sqrt {{{(x_t^i - \hat x_t^i)}^2} + {{(y_t^i - \hat y_t^i)}^2}} } } } }}{{(N + M){t_{pred}}}}, \hfill \\
  FDE = \frac{{\sum\nolimits_{i = 1}^N {\sum\nolimits_{m = 1}^M {\sqrt {{{(x_{t_{pred}}^i - \hat x_{t_{pred}}^i)}^2} + {{(y_{t_{pred}}^i - \hat y_{t_{pred}}^i)}^2}} } } }}{{(N + M){t_{pred}}}}, \hfill \\
\end{gathered}
\end{equation}
where $(x_i,y_i)$ and $(\hat x_i, \hat y_i)$ are the locations of the $i^{th}$ observed point (ground-truth) and its predicted one, $N$ is the number of trajectories, $M$ is the number of batches after partitioning the trajectory into some equal fragments, $t_{obs}$ and $t_{pred}$ are the numbers of observed frames and the ones to be predicted, respectively.

\textbf{Controlled experiments:} Similarly, we have conducted three groups of of experiments on location based prediction: 1) predicting future $5$ frames with observed $5$ frames (five-five prediction), 2) predicting future $10$ frames with observed $10$ frames (ten-ten prediction), and 3) predicting future $5$ frames with observed $10$ frames (ten-tive prediction), where the location of each point is the center of 3D bounding box in 3D point cloud. Consequently, the trajectories with over $10$ frames are valid for the first setting, and $20$ frames for the second one. The controlled experiments follow the Long Short-Term Memory (LSTM) network \cite{Hochreiter1997Long,Alahi2016Social}. The experimental details are as follows.

We partition the trajectories as many batches with the total number of observed frames and the ones to be predicted, and then pass the batches to train a LSTM network. The training and testing sets inherit the statistics of Table. I. We first generate the vectors with fixed length (set as $32$) by embedding the location of each annotation using a linear layer with relu nonlinearity. These embeddings are utilized as the input to the LSTM cell. The hidden state of LSTM is set as $64$, and the output of LSTM network utilizes a linear transform of $64$ to $2$ (location coordinates). We trained the network in $10$ epoch with the learning rate of $0.001$. The experimental results are demonstrated in Table IV. We can see that longer prediction cause larger predicted error, and longer observation will generate more accurate prediction.

\begin{table}\scriptsize
\centering
	\begin{threeparttable}
		\caption{The results of five-five prediction, ten-ten prediction and ten-five prediction. \emph{NTrainT} is the number of training trajectories, and \emph{NTestT} denotes the number of testing trajectories.}
		\label{tab:performance_comparison}
		\begin{tabular}{c|ccc|cc}
			\toprule
          &Classes &NTrainT&NTestT&ADE/meters&FDE/meters\cr
          \hline
           \multirow{3}{*}{Five-Five} &Pedestrians&160&157&0.34&0.49\cr
            &Vehicles&1,739&1,732&0.47&0.69\cr
            &Riders&563&557&0.44&0.66\cr
            \hline
            \multirow{3}{*}{Ten-Ten} &Pedestrians&122&94&0.44&0.79\cr
            &Vehicles&1,220&1,197&0.6&1.12\cr
            &Riders&344&357&0.65&1.26\cr
            \hline
            \multirow{3}{*}{Ten-Five} &Pedestrians&137&129&\textbf{0.27}&\textbf{0.39}\cr
            &Vehicles&1,450&1,456&\textbf{0.38}&\textbf{0.57}\cr
            &Riders&441&447&\textbf{0.38}&\textbf{0.56}\cr
			\bottomrule
		\end{tabular}
	\end{threeparttable}
\label{tab3}
\end{table}

\textbf{Metrics for 5D intention prediction:} Actually, in our benchmark, we provide not only the location prediction task, the event state chain, geometrical structure of 3D bounding boxes, and orientations should all be considered in prediction. Therefore, beside ADE and FDE in \cite{Alahi2016Social}. The precision and recall evaluation for predicted event state (stated in the metrics of 5D interactive event recognition), average precision (AP) of predicted 3D bounding boxes and orientation are advocated. Following ADE and FDE, we evaluate the prediction of 3D boxes and orientation as:
\begin{equation}\scriptsize
\begin{gathered}
\begin{gathered}
  AD{E_{orientation}} = \frac{{\sum\nolimits_{i = 1}^N {\sum\nolimits_{m = 1}^M {\sum\nolimits_{t = {t_{obs}} + 1}^{{t_{obs}} + {t_{pred}}} {\delta _t^i\left( {1 + \cos (\theta _t^i - \hat \theta _t^i)} \right)/2} } } }}{{(N + M){t_{pred}}}}, \hfill \\
  AD{E_{3Dbox}} = \frac{{\sum\nolimits_{i = 1}^N {\sum\nolimits_{m = 1}^M {\sum\nolimits_{t = {t_{obs}} + 1}^{{t_{obs}} + {t_{pred}}} {s(V_t^i,\hat V_t^i)} } } }}{{(N + M){t_{pred}}}}, \hfill \\
\end{gathered}
\end{gathered}
\end{equation}
 and
 \begin{equation}\scriptsize
\begin{gathered}
  FD{E_{orientation}} = \frac{{\sum\nolimits_{i = 1}^N {\sum\nolimits_{m = 1}^M {\delta _t^i\left( {1 + \cos (\theta _{{t_{pred}}}^i - \hat \theta _{{t_{pred}}}^i)} \right)/2} } }}{{(N + M){t_{pred}}}}, \hfill \\
  FD{E_{3Dbox}} = \frac{{\sum\nolimits_{i = 1}^N {\sum\nolimits_{m = 1}^M {s(V_{{t_{pred}}}^i,\hat V_{{t_{pred}}}^i)} } }}{{(N + M){t_{pred}}}}, \hfill \\
\end{gathered}
 \end{equation}
where $\theta$ and $\hat \theta$ are the observed orientations of targets and their predicted ones, and $s(V,\hat V)$ computes the overlapping rate of two 3D bounding boxes $V$ and the predicted $\hat V$ by PASCAL VOC criterion ~\cite{Everingham2010The}. $\delta$ penalizes the mis-predictions. If the predicted 3D bounding box overlaps the ground truth at least 50\%, $\delta=1$ and $0$ vice versa. In these metrics, $AD{E_{3Dbox}}$ and $FD{E_{3Dbox}}$ enjoy a larger value for better performance, and others pursuit smaller value. Therefore, our benchmark provide a platform for more challenging intention prediction task.

\section{CONCLUSIONS}

In this paper, we built a large-scale 5D semantics benchmark for autonomous driving which was captured under a wide range of interesting scenarios, and calibrated, synchronized and rectified efficiently and accurately. Different from the previously static detection/segmentation tasks, we focused on the deeper understanding of traffic scenes. Specifically, the tasks of 4D tracking, 5D interactive event recognition, and 5D intention prediction were launched in this benchmark. With the careful annotation, the benchmark yielded $249,129$ 3D annotations, $4,902$ independent instances for tracking with the length of overall $214,922$ points, $6,004$ 3D annotations for 5D interactive event recognition, and $4,900$ individuals for 5D intention prediction. These annotations were gathered under different light conditions (daytime and nighttime), diverse density of participants (low density and high density) and distinct driving scenarios (highway and urban). We believe that this benchmark will be highly useful in robotics and computer vision fields. In the future, we will embrace the 3D detection task, and make the task flow as an integrated chain, where each task can promote following ones. In addition, we will balance the annotations for a better utilization.

{\small
\bibliographystyle{IEEEtran}
\bibliography{egbib}
}

\end{document}